\documentclass{article}
\usepackage[final, nonatbib]{nips_2016}
\usepackage[english]{babel}
\usepackage[T1]{fontenc}
\usepackage{hyperref}
\usepackage{subcaption}
\usepackage{adjustbox}
\usepackage{graphicx}
\usepackage{xcolor}
\usepackage{amssymb}
\usepackage{amsmath}
\usepackage{csquotes}
\usepackage{lipsum}
\usepackage{setspace}
\usepackage{adjustbox}
\usepackage[square,numbers]{natbib}

\begin{document}

\title{Iterative Multi-document Neural Attention for Multiple Answer Prediction}

\author{
  Claudio Greco\\
  Department of Computer Science\\
  University of Bari Aldo Moro\\
  \texttt{claudiogaetanogreco@gmail.com} \\
  \And
  Alessandro Suglia\\
  Department of Computer Science\\
  University of Bari Aldo Moro\\
  \texttt{alessandro.suglia@gmail.com} \\
  \And
  Pierpaolo Basile\\
  Department of Computer Science\\
  University of Bari Aldo Moro\\
  \texttt{pierpaolo.basile@uniba.it} \\
  \And
  Gaetano Rossiello\\
  Department of Computer Science\\
  University of Bari Aldo Moro\\
  \texttt{gaetano.rossiello@uniba.it} \\
  \And
  Giovanni Semeraro\\
  Department of Computer Science\\
  University of Bari Aldo Moro\\
  \texttt{giovanni.semeraro@uniba.it} \\
}

\maketitle

\begin{abstract}
People have information needs of varying complexity, which can be solved by an intelligent agent able to answer questions formulated in a proper way, eventually considering user context and preferences. In a scenario in which the user profile can be considered as a question, intelligent agents able to answer questions can be used to find the most relevant answers for a given user. In this work we propose a novel model based on \textit{Artificial Neural Networks} to answer questions with multiple answers by exploiting multiple facts retrieved from a knowledge base. The model is evaluated on the \textit{factoid Question Answering} and \textit{top-n recommendation} tasks of the \textit{bAbI Movie Dialog} dataset. After assessing the performance of the model on both tasks, we try to define the long-term goal of a \textit{conversational recommender system} able to interact using natural language and to support users in their information seeking processes in a personalized way.
\end{abstract}

\section{Motivation and Background}
We are surrounded by a huge variety of technological artifacts which ``live'' with us today. These artifacts can help us in several ways because they have the power to accomplish complex and time-consuming tasks. Unfortunately, common software systems can do for us only specific types of tasks, in a strictly algorithmic way which is pre-defined by the software designer. \textit{Machine Learning (ML)}, a branch of \textit{Artificial Intelligence (AI)}, gives machines the ability to learn to complete tasks without being explicitly programmed.

People have information needs of varying complexity, ranging from simple questions about common facts which can be found in encyclopedias, to more sophisticated cases in which they need to know what movie to watch during a romantic evening. These tasks can be solved by an intelligent agent able to answer questions formulated in a proper way, eventually considering user context and preferences.

\textit{Question Answering (QA)} emerged in the last decade as one of the most promising fields in AI, since it allows to design intelligent systems which are able to give correct answers to user questions expressed in natural language. Whereas, \textit{recommender systems} produce individualized recommendations as output and have the effect of guiding the user in a personalized way to interesting or useful objects in a large space of possible options.
In a scenario in which the user profile (the set of user preferences) can be represented by a question, intelligent agents able to answer questions can be used to find the most appealing items for a given user, which is the classical task that recommender systems can solve. Despite the efficacy of classical recommender systems, generally they are not able to handle a conversation with the user so they miss the possibility of understanding his contextual information, emotions and feedback to refine the user profile and provide enhanced suggestions. \textit{Conversational recommender systems} assist online users in their information-seeking and decision making tasks by supporting an interactive process \cite{mahmood2009:crs} which could be goal oriented with the task of starting general and, through a series of interaction cycles, narrowing down the user interests until the desired item is obtained \cite{rubens2015:al}.

In this work we propose a novel model based on \textit{Artificial Neural Networks} to answer questions exploiting multiple facts retrieved from a knowledge base and evaluate it on a \textit{QA} task. Moreover, the effectiveness of the model is evaluated on the \textit{top-n recommendation} task, where the aim of the system is to produce a list of suggestions ranked according to the user preferences. After having assessed the performance of the model on both tasks, we try to define the long-term goal of a \textit{conversational recommender system} able to interact with the user using natural language and to support him in the information seeking process in a personalized way.

In order to fulfill our long-term goal of building a \textit{conversational recommender system} we need to assess the performance of our model on specific tasks involved in this scenario. A recent work which goes in this direction is reported in \cite{dodge2015:evaluating}, which presents the \textit{bAbI Movie Dialog} dataset, composed by different tasks such as \textit{factoid QA}, \textit{top-n recommendation} and two more complex tasks, one which mixes \textit{QA} and recommendation and one which contains turns of dialogs taken from Reddit. Having more specific tasks like \textit{QA} and recommendation, and a more complex one which mixes both tasks gives us the possibility to evaluate our model on different levels of granularity. Moreover, the subdivision in turns of the more complex task provides a proper benchmark of the model capability to handle an effective dialog with the user.

For the task related to \textit{QA}, a lot of datasets have been released in order to assess the machine reading and comprehension capabilities and a lot of neural network-based models have been proposed. Our model takes inspiration from \cite{sordoni2016:iterative}, which is able to answer \textit{Cloze-style} \cite{taylor1953:cloze} questions repeating an attention mechanism over the query and the documents multiple times. Despite the effectiveness on the \textit{Cloze-style} task, the original model does not consider multiple documents as a source of information to answer questions, which is fundamental in order to extract the answer from different relevant facts. The restricted assumption that the answer is contained in the given document does not allow the model to provide an answer which does not belong to the document. Moreover, this kind of task does not expect multiple answers for a given question, which is important for the complex information needs required for a \textit{conversational recommender system}.

According to our vision, the main outcomes of our work can be considered as building blocks for a \textit{conversational recommender system} and can be summarized as follows:
\begin{enumerate}
 \item we extend the model reported in \cite{sordoni2016:iterative} to let the inference process exploit evidences observed in multiple documents coming from an external knowledge base represented as a collection of textual documents;
 \item we design a model able to leverage the attention weights generated by the inference process to provide multiple answers which does not necessarily belong to the documents through a multi-layer neural network which may uncover possible relationships between the most relevant evidences;
 \item we assess the efficacy of our model through an experimental evaluation on \textit{factoid QA} and \textit{top-n recommendation} tasks supporting our hypothesis that a \textit{QA} model can be used to solve \textit{top-n recommendation}, too.
\end{enumerate}

The paper is organized as follows: Section \ref{sec:methodology} describes our model, while Section \ref{sec:evaluation} summarizes the evaluation of the model on the two above-mentioned tasks and the comparison with respect to state-of-the-art approaches. Section \ref{sec:related} gives an overview of the literature of both \textit{QA} and recommender systems, while final remarks and our long-term vision are reported in Section \ref{sec:concl_future}.

\section{Methodology}\label{sec:methodology}
Given a query $q$, an operator $\psi: Q \rightarrow D$ that produces the set of documents relevant for $q$, where $Q$ is the set of all queries and $D$ is the set of all documents. Our model defines a workflow in which a sequence of inference steps are performed in order to extract relevant information from $\psi(q)$ to generate the answers for $q$.

Following \cite{sordoni2016:iterative}, our workflow consists of three steps: (1) the \textit{encoding} phase, which generates meaningful representations for query and documents; (2) the \textit{inference} phase, which extracts relevant semantic relationships between the query and the documents by using an iterative attention mechanism and finally (3) the \textit{prediction} phase, which generates a score for each candidate answer.

\subsection{Encoding phase}
The input of the encoding phase is given by a query $q$ and a set of documents $\psi(q) = \{d_1, d_2, \dots, d_{|D_q|}\} \equiv D_q$. Both queries and documents are represented by a sequence of words $X = (x_1, x_2, \dots, x_{|X|})$, drawn from a vocabulary $V$. Each word is represented by a continuous $d$-dimensional word embedding $\mathbf{x} \in \mathbb{R}^d$ stored in a word embedding matrix $\mathbf{X} \in \mathbb{R}^{|V| \times d}$.

The sequences of dense representations for $q$ and $d_j$ are encoded using a \textit{bidirectional recurrent neural network encoder} with \textit{Gated Recurrent Units (GRU)} as in \cite{sordoni2016:iterative} which represents each word $x_i \in X$ as the concatenation of a forward encoding $\overrightarrow{\mathbf{h}_k} \in \mathbb{R}^{h}$ and a backward encoding $\overleftarrow{\mathbf{h}_k} \in \mathbb{R}^{h}$. From now on, we denote the contextual representation for the word $q_i$ by $\tilde{\mathbf{q}}_i \in \mathbb{R}^{2h}$ and the contextual representation for the word $d_{j, i}$ in the document $d_j$ by $\tilde{\mathbf{d}}_{j, i} \in \mathbb{R}^{2h}$. Differently from \cite{sordoni2016:iterative}, we build a unique representation for the whole set of documents $D_q$ related to the query $q$ by stacking each contextual representation $\tilde{\mathbf{d}}_{j, i}$ obtaining a matrix $\tilde{\mathbf{D}}_q \in \mathbb{R}^{l \times 2h}$, where $l = |d_1| + |d_2| + \dotsc + |d_{|D_q|}|$.

\subsection{Inference phase}
This phase uncovers a possible inference chain which models meaningful relationships between the query and the set of related documents. The inference chain is obtained by performing, for each inference step $t = 1, 2, \dots, T$, the attention mechanisms given by the \textit{query attentive read} and the \textit{document attentive read} keeping a state of the inference process given by an additional \textit{recurrent neural network} with \textit{GRU} units. In this way, the network is able to progressively refine the attention weights focusing on the most relevant tokens of the query and the documents which are exploited by the prediction neural network to select the correct answers among the candidate ones.

  \subsubsection{Query attentive read}
  Given the contextual representations for the query words $(\tilde{\mathbf{q}}_1, \tilde{\mathbf{q}}_2, \dots, \tilde{\mathbf{q}}_{|q|})$ and the inference \textit{GRU} state $\mathbf{s}_{t-1} \in \mathbb{R}^s$, we obtain a refined query representation $\mathbf{q}_t$ (\textit{query glimpse}) by performing an attention mechanism over the query at inference step $t$:
  \begin{gather*}
    \mathit{\hat{q}}_{i,t} = \underset{i=1, \dots, |q|}{\text{softmax }} \tilde{\mathbf{q}}_i^\top (\mathbf{A}_q \mathbf{s}_{t-1} + \mathbf{a}_q), \\
    \mathbf{q}_t = \sum_i \mathit{\hat{q}}_{i,t} \tilde{\mathbf{q}}_i
  \end{gather*}
  where $\hat{q}_{i, t}$ are the attention weights associated to the query words, $\mathbf{A}_q \in \mathbb{R}^{2h \times s}$ and $\mathbf{a}_q \in \mathbb{R}^{2h}$ are respectively a weight matrix and a bias vector which are used to perform the bilinear product with the query token representations $\tilde{\mathbf{q}}_i$. The attention weights can be interpreted as the relevance scores for each word of the query dependent on the inference state $s_{t-1}$ at the current inference step $t$.

  \subsubsection{Document attentive read}
  Given the query glimpse $\mathbf{q}_t$ and the inference \textit{GRU} state $\mathbf{s}_{t-1} \in \mathbb{R}^s$, we perform an attention mechanism over the contextual representations for the words of the stacked documents $\tilde{\mathbf{D}}_q$:
  \begin{gather*}
    \mathit{\hat{d}}_{i, t} = \underset{i=1, \dots, l}{\text{softmax }} \tilde{\mathbf{D}}_{q_i}^\top (\mathbf{A}_d [\mathbf{s}_{t-1}, \mathbf{q}_t] + \mathbf{a}_d), \\
    \mathbf{d}_t = \sum_i \mathit{\hat{d}}_{i, t} \tilde{\mathbf{D}}_{q_i}
  \end{gather*}
  where $\tilde{\mathbf{D}}_{q_i}$ is the $i$-th row of $\tilde{\mathbf{D}}_q$, $\hat{d}_{i, t}$ are the attention weights associated to the document words, $\mathbf{A}_d \in \mathbb{R}^{2h \times s}$ and $\mathbf{a}_d \in \mathbb{R}^{2h}$ are respectively a weight matrix and a bias vector which are used to perform the bilinear product with the document token representations $\tilde{\mathbf{D}}_{q_i}$. The attention weights can be interpreted as the relevance scores for each word of the documents conditioned on both the query glimpse and the inference state $s_{t-1}$ at the current inference step $t$. By combining the set of relevant documents in $\tilde{\mathbf{D}}_q$, we obtain the probability distribution ($\hat{d}_{1, t}, \hat{d}_{2, t}, \dots \hat{d}_{l, t}$) over all the relevant document tokens using the above-mentioned attention mechanism.
  
  \subsubsection{Gating search results}
  The inference \textit{GRU} state at the inference step $t$ is updated according to $\mathbf{s}_t = GRU([\mathbf{r}_q \cdot \mathbf{q}_t, \mathbf{r}_d \cdot \mathbf{d}_t], \mathbf{s}_{t-1})$, where $r_q$ and $r_d$ are the results of a gating mechanism obtained by evaluating $g([\mathbf{s}_{t-1}, \mathbf{q}_t, \mathbf{d}_t, \mathbf{q}_t \cdot \mathbf{d}_t])$ for the query and the documents, respectively. The gating function $g: \mathbb{R}^{s+6h} \rightarrow \mathbb{R}^{2h}$ is defined as a $2$-layer feed-forward neural network with a \textit{Rectified Linear Unit (ReLU)} \cite{nair2010:relu} activation function in the hidden layer and a \textit{sigmoid} activation function in the output layer. The purpose of the gating mechanism is to retain useful information for the inference process about query and documents and forget useless one.
  
\subsection{Prediction phase}
The prediction phase, which is completely different from the \textit{pointer-sum} loss reported in \cite{sordoni2016:iterative}, is able to generate, given the query $q$, a relevance score for each candidate answer $a \in A$ by using the document attention weights $\hat{d}_{i, T}$ computed in the last inference step $T$. The relevance score of each word $w$ is obtained by summing the attention weights of $w$ in each document related to $q$. Formally the relevance score for a given word $w$ is defined as:
\begin{equation*}
  score(w) = \frac{1}{\pi(w)} \sum\limits_{i=1}^{l} \phi(i, w)
\end{equation*}
where $\phi(i, w)$ returns $0$ if $\sigma(i) \neq w$, $\hat{d}_{i, T}$ otherwise; $\sigma(i)$ returns the word in position $i$ of the stacked documents matrix $\tilde{\mathbf{D}}_q$ and $\pi(w)$ returns the frequency of the word $w$ in the documents $D_q$ related to the query $q$. The relevance score takes into account the importance of token occurrences in the considered documents given by the computed attention weights. Moreover, the normalization term $\frac{1}{\pi(w)}$ is applied to the relevance score in order to mitigate the weight associated to highly frequent tokens.

The evaluated relevance scores are concatenated in a single vector representation $\mathbf{z} = [score(w_1), score(w_2), \dots, score(w_{|V|})]$ which is given in input to the answer prediction neural network defined as:
\begin{gather*}
 \mathbf{y} = \text{sigmoid} (\mathbf{W}_{ho} \text{ relu} (\mathbf{W}_{ih} \mathbf{z} + \mathbf{b}_{ih}) + \mathbf{b}_{ho})
\end{gather*}
where $u$ is the hidden layer size, $\mathbf{W}_{ih} \in \mathbb{R}^{u \times |V|}$ and $\mathbf{W}_{ho} \in \mathbb{R}^{|A| \times u}$ are weight matrices, $\mathbf{b}_{ih} \in \mathbb{R}^{u}$, $\mathbf{b}_{ho} \in \mathbb{R}^{|A|}$ are bias vectors, $\text{sigmoid(x)} = \frac{1}{1 + e^{-x}}$ is the \textit{sigmoid} function and $\text{relu(x)} = max(0, x)$ is the \textit{ReLU} activation function, which are applied pointwise to the given input vector.

The neural network weights are supposed to learn latent features which encode relationships between the most relevant words for the given query to predict the correct answers. The outer \textit{sigmoid} activation function is used to treat the problem as a \textit{multi-label} classification problem, so that each candidate answer is independent and not mutually exclusive. In this way the neural network generates a score which represents the probability that the candidate answer is correct. Moreover, differently from \cite{sordoni2016:iterative}, the candidate answer $A$ can be any word, even those which not belong to the documents related to the query. 

The model is trained by minimizing the \textit{binary cross-entropy} loss function comparing the neural network output $y$ with the target answers for the given query $q$ represented as a binary vector, in which there is a $1$ in the corresponding position of the correct answer, $0$ otherwise.

\section{Experimental evaluation}\label{sec:evaluation}
The model performance is evaluated on the \textit{QA} and \textit{Recs} tasks of the \textit{bAbI Movie Dialog} dataset using \textit{HITS@k} evaluation metric, which is equal to the number of correct answers in the top-$k$ results. In particular, the performance for the \textit{QA} task is evaluated according to \textit{HITS@1}, while the performance for the \textit{Recs} task is evaluated according to \textit{HITS@100}.

Differently from \cite{dodge2015:evaluating}, the relevant knowledge base facts, taken from the knowledge base in triple form distributed with the dataset, are retrieved by $\psi$ implemented by exploiting the \textit{Elasticsearch} engine and not according to an hash lookup operator which applies a strict filtering procedure based on word frequency. In our work, $\psi$ returns at most the top $30$ relevant facts for $q$. Each entity in questions and documents is recognized using the list of entities provided with the dataset and considered as a single word of the dictionary $V$.

Questions, answers and documents given in input to the model are preprocessed using the \textit{NLTK} toolkit \cite{bird2006:nltk} performing only word tokenization. The question given in input to the $\psi$ operator is preprocessed performing word tokenization and stopword removal.

The optimization method and tricks are adopted from \cite{sordoni2016:iterative}. The model is trained using \textit{ADAM} \cite{kingma2014:adam} optimizer (\textit{learning rate}=$0.001$) with a batch size of $128$ for at most $100$ epochs considering the best model until the \textit{HITS@k} on the validation set decreases for $5$ consecutive times. \textit{Dropout} \cite{srivastava2014:dropout} is applied on $r_q$ and on $r_d$ with a rate of $0.2$ and on the prediction neural network hidden layer with a rate of $0.5$. \textit{L2 regularization} is applied to the embedding matrix $\mathbf{X}$ with a coefficient equal to $0.0001$. We clipped the gradients if their norm is greater than $5$ to stabilize learning \cite{pascanu2013:difficulty}. Embedding size $d$ is fixed to $50$. All \textit{GRU} output sizes are fixed to $128$. The number of inference steps $T$ is set to $3$. The size of the prediction neural network hidden layer $u$ is fixed to $4096$. Biases $\mathbf{b}_{ih}$ and $\mathbf{b}_{ho}$ are initialized to zero vectors. All weight matrices are initialized sampling from the normal distribution $\mathcal{N}(0, 0.05)$. The \textit{ReLU} activation function in the prediction neural network has been experimentally chosen comparing different activation functions such as \textit{sigmoid} and \textit{tanh} and taking the one which leads to the best performance. The model is implemented in \textit{TensorFlow} \cite{abadi2016:tensorflow} and executed on an \textit{NVIDIA TITAN X} GPU.

\begin{table}[!ht]
\centering
\small
\begin{tabular}{|l|l|l|}
\hline
\textbf{METHODS} & \textbf{QA TASK} & \textbf{RECS TASK} \\ \hline
QA SYSTEM                                      & \textbf{90.7}                                                                                             & N/A                                                                                                  \\
SVD                                            & N/A                                                                                              & 19.2                                                                                                                                                                                                 \\ \hline
LSTM                                           & 6.5                                                                                              & 27.1                                                                                                 \\
SUPERVISED EMBEDDINGS                          & 50.9                                                                                             & 29.2                                                                                                 \\
MEMN2N                                         & 79.3                                                                                             & 28.6                                                                                                 \\ \hline\hline
JOINT SUPERVISED EMBEDDINGS                    & 43.6                                                                                             & 28.1                                                                                                 \\
JOINT MEMN2N                                   & 83.5                                                                                             & 26.5                                                                                                 \\ \hline\hline
OUR MODEL                                           & 86.8                                                                                    &   \textbf{30.0}                                                                                                  \\ \hline
\end{tabular}
\caption{Comparison between our model and baselines from \cite{dodge2015:evaluating} on the \textit{QA} and \textit{Recs} tasks evaluated according to \textit{HITS@1} and \textit{HITS@100}, respectively.}
\label{table:results}
\end{table}

Following the experimental design, the results in Table \ref{table:results} are promising because our model outperforms all other systems on both tasks except for the \textit{QA SYSTEM} on the \textit{QA} task. Despite the advantage of the \textit{QA SYSTEM}, it is a carefully designed system to handle knowledge base data in the form of triples, but our model can leverage data in the form of documents, without making any assumption about the form of the input data and can be applied to different kind of tasks. Additionally, the model \textit{MEMN2N} is a neural network whose weights are pre-trained on the same dataset without using the long-term memory and the models \textit{JOINT SUPERVISED EMBEDDINGS} and \textit{JOINT MEMN2N} are models trained across all the tasks of the dataset in order to boost performance. Despite that, our model outperforms the three above-mentioned ones without using any supplementary trick.
Even though our model performance is higher than all the others on the \textit{Recs} task, we believe that the obtained result may be improved and so we plan a further investigation. Moreover, the need for further investigation can be justified by the work reported in \cite{searle2016:blow_out} which describes some issues regarding the \textit{Recs} task.

\begin{figure}[!ht]
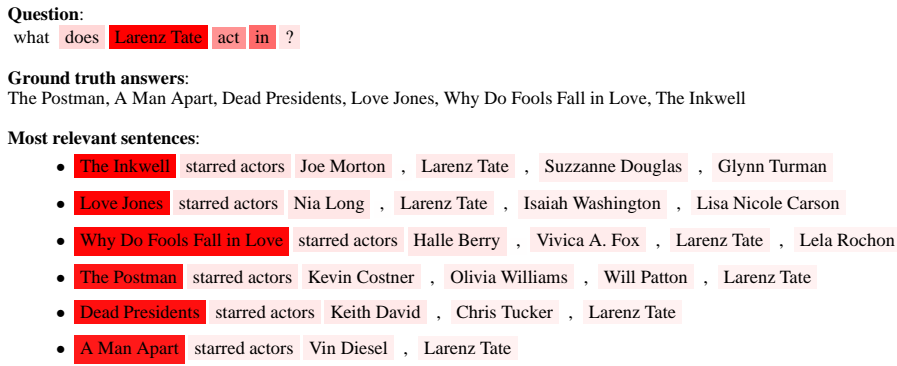

    \centering
    \adjustbox{minipage=[r][0.285\textheight][b]{0.85\paperwidth},scale={0.7}}{
    \textbf{Question}:\\\colorbox{red!0.0}{what} \colorbox{red!15.848116565744633}{does} \colorbox{red!100.0}{Larenz Tate} \colorbox{red!42.330914370449236}{act} \colorbox{red!58.45324143686287}{in} \colorbox{red!10.14420812774938}{?}\\
    
    \textbf{Ground truth answers}:\\ The Postman, A Man Apart, Dead Presidents, Love Jones, Why Do Fools Fall in Love, The Inkwell\\
    
    \textbf{Most relevant sentences}:
    \begin{itemize}
     \item \colorbox{red!100.0}{The Inkwell} \colorbox{red!11.014547069535482}{starred actors} \colorbox{red!10.643648669028279}{Joe Morton} \colorbox{red!0.3408869667466786}{,} \colorbox{red!8.134645882386382}{Larenz Tate} \colorbox{red!0.19150103305571511}{,} \colorbox{red!6.24307758889328}{Suzzanne Douglas} \colorbox{red!0.02451639598545446}{,} \colorbox{red!5.190426688487163}{Glynn Turman}
     \item \colorbox{red!98.28247096240496}{Love Jones} \colorbox{red!10.0595480081499}{starred actors} \colorbox{red!9.590697315557215}{Nia Long} \colorbox{red!0.3056431045858451}{,} \colorbox{red!7.594042404971506}{Larenz Tate} \colorbox{red!0.1609306034972888}{,} \colorbox{red!5.795733977352118}{Isaiah Washington} \colorbox{red!0.005091172401225019}{,} \colorbox{red!4.815556949124064}{Lisa Nicole Carson}
     \item \colorbox{red!97.20994002041111}{Why Do Fools Fall in Love} \colorbox{red!9.566514097142505}{starred actors} \colorbox{red!8.934382282090624}{Halle Berry} \colorbox{red!0.29334246998983643}{,} \colorbox{red!7.543080972243575}{Vivica A. Fox} \colorbox{red!0.15508707844115807}{,} \colorbox{red!5.786286355116288}{Larenz Tate} \colorbox{red!0.0}{,} \colorbox{red!4.7300858983734715}{Lela Rochon}
     \item \colorbox{red!91.80880282259236}{The Postman} \colorbox{red!9.103530085542806}{starred actors} \colorbox{red!7.709442408418296}{Kevin Costner} \colorbox{red!0.24065625486630504}{,} \colorbox{red!6.915380552842576}{Olivia Williams} \colorbox{red!0.16902078069940646}{,} \colorbox{red!6.084254654078019}{Will Patton} \colorbox{red!0.012201480300291967}{,} \colorbox{red!5.02143695779399}{Larenz Tate}
     \item \colorbox{red!93.97061352966541}{Dead Presidents} \colorbox{red!8.763482282292808}{starred actors} \colorbox{red!8.833058026308105}{Keith David} \colorbox{red!0.16423070724153135}{,} \colorbox{red!5.749700740374816}{Chris Tucker} \colorbox{red!0.008776123988528943}{,} \colorbox{red!5.116757893418958}{Larenz Tate}
     \item \colorbox{red!90.18000274210308}{A Man Apart} \colorbox{red!6.622857917840147}{starred actors} \colorbox{red!6.837951692920831}{Vin Diesel} \colorbox{red!0.00982663325183327}{,} \colorbox{red!5.148639599665926}{Larenz Tate}
    \end{itemize}
    }
    \caption{Attention weights $\tilde{\mathbf{q}}_i$ and $\tilde{\mathbf{D}}_{q_i}$ computed by the neural network attention mechanisms at the last inference step $T$ for each token. Higher shades correspond to higher relevance scores for the related tokens.}
    \label{fig:attention_weights}
\end{figure}

Figure \ref{fig:attention_weights} shows the attention weights computed in the last inference step of the iterative attention mechanism used by the model to answer to a given question. Attention weights, represented as red boxes with variable color shades around the tokens, can be used to interpret the reasoning mechanism applied by the model because higher shades of red are associated to more relevant tokens on which the model focus its attention. It is worth to notice that the attention weights associated to each token are the result of the inference mechanism uncovered by the model which progressively tries to focus on the relevant aspects of the query and the documents which are exploited to generate the answers.

Given the question ``what does Larenz Tate act in?'' shown in the above-mentioned figure, the model is able to understand that ``Larenz Tate'' is the subject of the question and ``act in'' represents the intent of the question. Reading the related documents, the model associates higher attention weights to the most relevant tokens needed to answer the question, such as ``The Postman'', ``A Man Apart'' and so on.

\section{Related work}\label{sec:related}

We think that it is necessary to consider models and techniques coming from research both in \textit{QA} and recommender systems in order to pursue our desire to build an intelligent agent able to assist the user in decision-making tasks. We cannot fill the gap between the above-mentioned research areas if we do not consider the proposed models in a synergic way by virtue of the proposed analogy between the user profile (the set of user preferences) and the items to be recommended, as the question and the correct answers. The first work which goes in this direction is reported in \cite{musto2016:amar}, which exploits movie descriptions to suggest appealing movies for a given user using an architecture tipically used for \textit{QA} tasks. In fact, most of the research in the recommender systems field presents ad-hoc systems which exploit neighbourhood information like in \textit{Collaborative Filtering} techniques \cite{ning2015:cf}, item descriptions and metadata like in \textit{Content-based} systems \cite{deg2015:scbrs}. Recently presented neural network models \cite{covingto2016:youtube, cheng2016:wide_deep} systems are able to learn latent representations in the network weights leveraging information coming from user preferences and item information.

In recent days, a lot of effort is devoted to create benchmarks for artificial agents to assess their ability to comprehend natural language and to reason over facts. One of the first attempt is the \textit{bAbI} \cite{weston2015:babi} dataset which is a synthetic dataset containing elementary tasks such as selecting an answer between one or more candidate facts, answering yes/no questions, counting operations over lists and sets and basic induction and deduction tasks. Another relevant benchmark is the one described in \cite{hermann2015:teaching}, which provides \textit{CNN/Daily Mail} datasets consisting of document-query-answer triples where an entity in the query is replaced by a placeholder and the system should identify the correct entity by reading and comprehending the given document. \textit{MCTest} \cite{richardson2013:mctest} requires machines to answer multiple-choice reading comprehension questions about fictional stories, directly tackling the high-level goal of open-domain machine comprehension. Finally, \textit{SQuAD} \cite{rajpurkar2016:squad} consists in a set of \textit{Wikipedia} articles, where the answer to each question is a segment of text from the corresponding reading passage.

According to the experimental evaluations conducted on the above-mentioned datasets, high-level performance can be obtained exploiting complex attention mechanisms which are able to focus on relevant evidences in the processed content. One of the earlier approaches used to solve these tasks is given by the general \textit{Memory Network} \cite{weston2014:memnn, sukhbaatar2015:e2e_memnn} framework which is one of the first neural network models able to access external memories to extract relevant information through an attention mechanism and to use them to provide the correct answer. A deep \textit{Recurrent Neural Network} with \textit{Long Short-Term Memory} units is presented in \cite{hermann2015:teaching}, which solves \textit{CNN/Daily Mail} datasets by designing two different attention mechanisms called \textit{Impatient Reader} and \textit{Attentive Reader}. Another way to incorporate attention in neural network models is proposed in \cite{kadlec2016:as_reader} which defines a \textit{pointer-sum} loss whose aim is to maximize the attention weights which lead to the correct answer.

\section{Conclusions and Future Work}\label{sec:concl_future}
In this work we propose a novel model based on \textit{Artificial Neural Networks} to answer questions with multiple answers by exploiting multiple facts retrieved from a knowledge base. The proposed model can be considered a relevant building block of a \textit{conversational recommender system}. Differently from \cite{sordoni2016:iterative}, our model can consider multiple documents as a source of information in order to generate multiple answers which may not belong to the documents.
As presented in this work, common tasks such as \textit{QA} and \textit{top-n recommendation} can be solved effectively by our model.

In a common recommendation system scenario, when a user enters a search query, it is assumed that his preferences are known. This is a stringent requirement because users cannot have a clear idea of their preferences at that point. Conversational recommender systems support users to fulfill their information needs through an interactive process. In this way, the system can provide a personalized experience dynamically adapting the user model with the possibility to enhance the generated predictions. Moreover, the system capability can be further enhanced giving explanations to the user about the given suggestions.

To reach our goal, we should improve our model by designing a $\psi$ operator able to return relevant facts recognizing the most relevant information in the query, by exploiting user preferences and contextual information to learn the user model and by providing a mechanism which leverages attention weights to give explanations. In order to effectively train our model, we plan to collect real dialog data containing contextual information associated to each user and feedback for each dialog which represents if the user is satisfied with the conversation. Given these enhancements, we should design a system able to hold effectively a dialog with the user recognizing his intent and providing him the most suitable contents.

With this work we try to show the effectiveness of our architecture for tasks which go from \emph{pure question answering} to \textit{top-n recommendation} through an experimental evaluation without any assumption on the task to be solved. To do that, we do not use any hand-crafted linguistic features but we let the system learn and leverage them in the inference process which leads to the answers through multiple reasoning steps. During these steps, the system understands relevant relationships between question and documents without relying on canonical matching, but repeating an attention mechanism able to unconver related aspects in distributed representations, conditioned on an encoding of the inference process given by another neural network. Equipping agents with a reasoning mechanism like the one described in this work and exploiting the ability of neural network models to learn from data, we may be able to create truly intelligent agents.

\section{Acknowledgments}
This work is supported by the \textit{IBM Faculty Award "Deep Learning to boost Cognitive Question Answering"}. The Titan X GPU used for this research was donated by the \textit{NVIDIA Corporation}.

\bibliographystyle{abbrvnat}
\bibliography{bibliography}

\begin{thebibliography}{24}
\providecommand{\natexlab}[1]{#1}
\providecommand{\url}[1]{\texttt{#1}}
\expandafter\ifx\csname urlstyle\endcsname\relax
  \providecommand{\doi}[1]{doi: #1}\else
  \providecommand{\doi}{doi: \begingroup \urlstyle{rm}\Url}\fi

\bibitem[Abadi et~al.(2016)Abadi, Agarwal, Barham, Brevdo, Chen, Citro,
  Corrado, Davis, Dean, Devin, Ghemawat, Goodfellow, Harp, Irving, Isard, Jia,
  J{\'{o}}zefowicz, Kaiser, Kudlur, Levenberg, Man{\'{e}}, Monga, Moore,
  Murray, Olah, Schuster, Shlens, Steiner, Sutskever, Talwar, Tucker,
  Vanhoucke, Vasudevan, Vi{\'{e}}gas, Vinyals, Warden, Wattenberg, Wicke, Yu,
  and Zheng]{abadi2016:tensorflow}
M.~Abadi, A.~Agarwal, P.~Barham, E.~Brevdo, Z.~Chen, C.~Citro, G.~S. Corrado,
  A.~Davis, J.~Dean, M.~Devin, S.~Ghemawat, I.~J. Goodfellow, A.~Harp,
  G.~Irving, M.~Isard, Y.~Jia, R.~J{\'{o}}zefowicz, L.~Kaiser, M.~Kudlur,
  J.~Levenberg, D.~Man{\'{e}}, R.~Monga, S.~Moore, D.~G. Murray, C.~Olah,
  M.~Schuster, J.~Shlens, B.~Steiner, I.~Sutskever, K.~Talwar, P.~A. Tucker,
  V.~Vanhoucke, V.~Vasudevan, F.~B. Vi{\'{e}}gas, O.~Vinyals, P.~Warden,
  M.~Wattenberg, M.~Wicke, Y.~Yu, and X.~Zheng.
\newblock Tensorflow: Large-scale machine learning on heterogeneous distributed
  systems.
\newblock \emph{CoRR}, abs/1603.04467, 2016.

\bibitem[Bird(2006)]{bird2006:nltk}
S.~Bird.
\newblock Nltk: the natural language toolkit.
\newblock In \emph{Proceedings of the COLING/ACL on Interactive presentation
  sessions}, pages 69--72. Association for Computational Linguistics, 2006.

\bibitem[Cheng et~al.(2016)Cheng, Koc, Harmsen, Shaked, Chandra, Aradhye,
  Anderson, Corrado, Chai, Ispir, Anil, Haque, Hong, Jain, Liu, and
  Shah]{cheng2016:wide_deep}
H.~Cheng, L.~Koc, J.~Harmsen, T.~Shaked, T.~Chandra, H.~Aradhye, G.~Anderson,
  G.~Corrado, W.~Chai, M.~Ispir, R.~Anil, Z.~Haque, L.~Hong, V.~Jain, X.~Liu,
  and H.~Shah.
\newblock Wide {\&} deep learning for recommender systems.
\newblock \emph{CoRR}, abs/1606.07792, 2016.

\bibitem[Covington et~al.(2016)Covington, Adams, and
  Sargin]{covingto2016:youtube}
P.~Covington, J.~Adams, and E.~Sargin.
\newblock Deep neural networks for youtube recommendations.
\newblock In \emph{Proceedings of the 10th ACM Conference on Recommender
  Systems}, New York, NY, USA, 2016.

\bibitem[de~Gemmis et~al.(2015)de~Gemmis, Lops, Musto, Narducci, and
  Semeraro]{deg2015:scbrs}
M.~de~Gemmis, P.~Lops, C.~Musto, F.~Narducci, and G.~Semeraro.
\newblock Semantics-aware content-based recommender systems.
\newblock In \emph{Recommender Systems Handbook}, pages 119--159. Springer,
  2015.

\bibitem[Dodge et~al.(2015)Dodge, Gane, Zhang, Bordes, Chopra, Miller, Szlam,
  and Weston]{dodge2015:evaluating}
J.~Dodge, A.~Gane, X.~Zhang, A.~Bordes, S.~Chopra, A.~Miller, A.~Szlam, and
  J.~Weston.
\newblock Evaluating prerequisite qualities for learning end-to-end dialog
  systems.
\newblock \emph{arXiv preprint arXiv:1511.06931}, 2015.

\bibitem[Hermann et~al.(2015)Hermann, Kocisk{\'{y}}, Grefenstette, Espeholt,
  Kay, Suleyman, and Blunsom]{hermann2015:teaching}
K.~M. Hermann, T.~Kocisk{\'{y}}, E.~Grefenstette, L.~Espeholt, W.~Kay,
  M.~Suleyman, and P.~Blunsom.
\newblock Teaching machines to read and comprehend.
\newblock In \emph{Advances in Neural Information Processing Systems}, pages
  1693--1701, 2015.

\bibitem[Kadlec et~al.(2016)Kadlec, Schmid, Bajgar, and
  Kleindienst]{kadlec2016:as_reader}
R.~Kadlec, M.~Schmid, O.~Bajgar, and J.~Kleindienst.
\newblock Text understanding with the attention sum reader network.
\newblock \emph{arXiv preprint arXiv:1603.01547}, 2016.

\bibitem[Kingma and Ba(2014)]{kingma2014:adam}
D.~Kingma and J.~Ba.
\newblock Adam: A method for stochastic optimization.
\newblock \emph{arXiv preprint arXiv:1412.6980}, 2014.

\bibitem[Mahmood and Ricci(2009)]{mahmood2009:crs}
T.~Mahmood and F.~Ricci.
\newblock Improving recommender systems with adaptive conversational
  strategies.
\newblock In \emph{Proceedings of the 20th ACM conference on Hypertext and
  hypermedia}, pages 73--82. ACM, 2009.

\bibitem[Musto et~al.()Musto, Greco, Suglia, and Semeraro]{musto2016:amar}
C.~Musto, C.~Greco, A.~Suglia, and G.~Semeraro.
\newblock Ask me any rating: {A} content-based recommender system based on
  recurrent neural networks.
\newblock In \emph{Proceedings of the 7th Italian Information Retrieval
  Workshop, Venezia, Italy, May 30-31, 2016.}

\bibitem[Nair and Hinton(2010)]{nair2010:relu}
V.~Nair and G.~E. Hinton.
\newblock Rectified linear units improve restricted boltzmann machines.
\newblock In \emph{Proceedings of the 27th International Conference on Machine
  Learning (ICML-10)}, pages 807--814, 2010.

\bibitem[Ning et~al.(2015)Ning, Desrosiers, and Karypis]{ning2015:cf}
X.~Ning, C.~Desrosiers, and G.~Karypis.
\newblock A comprehensive survey of neighborhood-based recommendation methods.
\newblock In \emph{Recommender Systems Handbook}, pages 37--76. Springer, 2015.

\bibitem[Pascanu et~al.(2013)Pascanu, Mikolov, and
  Bengio]{pascanu2013:difficulty}
R.~Pascanu, T.~Mikolov, and Y.~Bengio.
\newblock On the difficulty of training recurrent neural networks.
\newblock \emph{ICML (3)}, 28:\penalty0 1310--1318, 2013.

\bibitem[Rajpurkar et~al.(2016)Rajpurkar, Zhang, Lopyrev, and
  Liang]{rajpurkar2016:squad}
P.~Rajpurkar, J.~Zhang, K.~Lopyrev, and P.~Liang.
\newblock Squad: 100, 000+ questions for machine comprehension of text.
\newblock \emph{CoRR}, abs/1606.05250, 2016.

\bibitem[Richardson et~al.(2013)Richardson, Burges, and
  Renshaw]{richardson2013:mctest}
M.~Richardson, C.~J.~C. Burges, and E.~Renshaw.
\newblock Mctest: A challenge dataset for the open-domain machine comprehension
  of text.
\newblock In \emph{EMNLP}, 2013.

\bibitem[Rubens et~al.(2015)Rubens, Kaplan, and Sugiyama]{rubens2015:al}
N.~Rubens, D.~Kaplan, and M.~Sugiyama.
\newblock Active learning in recommender systems.
\newblock In \emph{Recommender Systems Handbook}, pages 809--846. Springer,
  2015.

\bibitem[Searle and Bingham-Walker(2016)]{searle2016:blow_out}
R.~Searle and M.~Bingham-Walker.
\newblock Why ``blow out''? a structural analysis of the movie dialog dataset.
\newblock \emph{ACL 2016}, page 215, 2016.

\bibitem[Sordoni et~al.(2016)Sordoni, Bachman, and
  Bengio]{sordoni2016:iterative}
A.~Sordoni, P.~Bachman, and Y.~Bengio.
\newblock Iterative alternating neural attention for machine reading.
\newblock \emph{arXiv preprint arXiv:1606.02245}, 2016.

\bibitem[Srivastava et~al.(2014)Srivastava, Hinton, Krizhevsky, Sutskever, and
  Salakhutdinov]{srivastava2014:dropout}
N.~Srivastava, G.~E. Hinton, A.~Krizhevsky, I.~Sutskever, and R.~Salakhutdinov.
\newblock Dropout: a simple way to prevent neural networks from overfitting.
\newblock \emph{Journal of Machine Learning Research}, 15\penalty0
  (1):\penalty0 1929--1958, 2014.

\bibitem[Sukhbaatar et~al.(2015)Sukhbaatar, Szlam, Weston, and
  Fergus]{sukhbaatar2015:e2e_memnn}
S.~Sukhbaatar, A.~Szlam, J.~Weston, and R.~Fergus.
\newblock End-to-end memory networks.
\newblock In \emph{Advances in neural information processing systems}, pages
  2440--2448, 2015.

\bibitem[Taylor(1953)]{taylor1953:cloze}
W.~L. Taylor.
\newblock Cloze procedure: a new tool for measuring readability.
\newblock \emph{Journalism and Mass Communication Quarterly}, 30\penalty0
  (4):\penalty0 415, 1953.

\bibitem[Weston et~al.(2014)Weston, Chopra, and Bordes]{weston2014:memnn}
J.~Weston, S.~Chopra, and A.~Bordes.
\newblock Memory networks.
\newblock \emph{arXiv preprint arXiv:1410.3916}, 2014.

\bibitem[Weston et~al.(2015)Weston, Bordes, Chopra, and
  Mikolov]{weston2015:babi}
J.~Weston, A.~Bordes, S.~Chopra, and T.~Mikolov.
\newblock Towards ai-complete question answering: {A} set of prerequisite toy
  tasks.
\newblock \emph{CoRR}, abs/1502.05698, 2015.

\end{thebibliography}
\end{document}